\pdfoutput=1

\documentclass[11pt]{article}

\usepackage[]{acl}

\usepackage[T1]{fontenc}

\usepackage[utf8]{inputenc}

\usepackage{microtype}
\usepackage{times}
\usepackage{latexsym}
\usepackage{amssymb}
\usepackage{amsfonts}
\usepackage{amsmath} 
\usepackage{amsthm} 
\usepackage{booktabs}
\usepackage{enumerate}
\usepackage{graphicx}
\usepackage{subfigure}
\usepackage{xspace}
\usepackage{float}
\usepackage{bbm}
\usepackage{bm}
\usepackage{multirow}
\usepackage{booktabs}
\usepackage{color}
\usepackage{framed}
\usepackage{stfloats}
\usepackage{iitem}
\usepackage{makecell}
\usepackage[normalem]{ulem}
\useunder{\uline}{\ul}{}
\usepackage{url}
\newcommand{\paratitle}[1]{\vspace{1.5ex}\noindent\textbf{#1}}
\newcommand{\ie}{\emph{i.e.,}\xspace}

\newcommand{\eg}{\emph{e.g.,}\xspace}

\newcommand{\ignore}[1]{}

\usepackage{xcolor}
\definecolor{tOrange}{RGB}{255,165,0}
\definecolor{tBlue}{RGB}{24,116,205}
\definecolor{tPink}{RGB}{255,20,147}
\definecolor{tGreen}{RGB}{50,205,50}
\definecolor{tGold}{RGB}{255,215,0}

\title{Learning to Transfer Prompts for Text Generation}

\author{
	Junyi Li\textsuperscript{\rm{1,3,4}}, 
	Tianyi Tang\textsuperscript{\rm{1}}, 
	Jian-Yun Nie\textsuperscript{\rm{3}}, 
	Ji-Rong Wen\textsuperscript{\rm{1,2,4}}
	{\rm and} Wayne Xin Zhao\textsuperscript{\rm{1,4}\thanks{\ \ Corresponding author}\ } \\
	\textsuperscript{1}Gaoling School of Artificial Intelligence, Renmin University of China \\
	\textsuperscript{2}School of Information, Renmin University of China \\
	\textsuperscript{3}DIRO, Universit\'{e} de Montr\'{e}al \\
	\textsuperscript{4}Beijing Key Laboratory of Big Data Management and Analysis Methods \\
	\texttt{\{lijunyi,steven\_tang,jrwen\}@ruc.edu.cn} \\ 
	\texttt{nie@iro.umontreal.ca} \quad
	\texttt{batmanfly@gmail.com} \\
}

\begin{document}
\maketitle

\begin{abstract}
Pretrained language models (PLMs) have made remarkable progress in text generation tasks via fine-tuning. While, it is challenging to fine-tune PLMs in a data-scarce situation. Therefore, it is non-trivial to develop a general and lightweight model that can adapt to various text generation tasks based on PLMs. To fulfill this purpose, the recent prompt-based learning offers a potential solution. In this paper, we improve this technique and propose a novel prompt-based method (\textsc{Ptg}) for text generation in a transferable setting. First, \textsc{Ptg} learns a set of source prompts for various source generation tasks and then transfers these prompts as target prompts to perform target generation tasks. To consider both task- and instance-level information, we design an adaptive attention mechanism to derive the target prompts. For each data instance, \textsc{Ptg} learns a specific target prompt by attending to highly relevant source prompts. In extensive experiments, \textsc{Ptg} yields competitive or better results than fine-tuning methods. We release our source prompts as an open resource, where users can add or reuse them to improve new text generation tasks for future research. Code and data can be available at \url{https://github.com/RUCAIBox/Transfer-Prompts-for-Text-Generation}.

\end{abstract}

\section{Introduction}
\label{sec-intro}

In natural language processing~(NLP), text generation is an important research topic that aims to automatically produce understandable text in human language from input data~\cite{PLM4TG_survey}. In recent decades, various approaches have been widely applied to a variety of text generation tasks~\cite{LiZWS19,GehringAGYD17,textbox}, especially the emergence of pretrained language models (PLMs)~\cite{LiTZW21}. By involving large-scale parameters pretrained on massive general corpora, PLMs such as GPT-3~\cite{gpt3} have achieved substantial progress in text generation. Through the \textit{fine-tuning} paradigm, PLMs can adapt to various text generation tasks by directly adjusting the model parameters with labelled datasets. 


However, in real-world scenarios, we are inevitably confronted with tasks having only limited labelled data (\eg new domains). It is often difficult to fine-tune text generation models in a data-scarce situation~\cite{ChenECLW20,LiTZWYW21}. Although the input and output formats are different for various text generation tasks, these tasks essentially adopt similar learning and generation mechanism (\eg Seq2Seq~\cite{SutskeverVL14}). Furthermore, the success of PLMs sheds light on the possibility of developing general or transferable text generation models. For example, \citet{radford2019language} framed generation tasks as language modeling by predicting the next token given previous tokens. Based on these studies, we aim to devise a general and lightweight text generation approach that can effectively adapt to various new tasks and datasets, based on PLMs.

To fulfill this purpose, the recently proposed \emph{prompt-based learning} offers a potential technical solution~\cite{prompt_survey}. In this paradigm, a text generation task can be solved with the help of a \textit{prompt} containing task-specific information. For example, T5~\cite{RaffelSRLNMZLL20} framed summarization and question answering into a text-to-text format by utilizing prompts ``\texttt{summarize:}'' and ``\texttt{answer the question:}''. Based on learned or manually designed prompts, PLMs can be leveraged  to perform existing or new generation tasks without being tuned~\cite{
gpt3,prefix}, which provides a unified approach to utilizing PLMs for various generation tasks. Furthermore, to quickly adapt PLMs to new NLU tasks, several works directly used a soft prompt learned from source NLU tasks to initialize the prompt for a target NLU task~\cite{spot,tpt}. Inspired by these studies, we aim to apply prompt-based methods to data-scarce text generation tasks in a transferable setting.


Despite promising, there are still two major challenges for transferring prompts in text generation.
Firstly, it has been found that prompts are highly task-specific~\cite{gao2020making}, and it is difficult to effectively transfer or reuse existing prompts for new tasks. Second, for a single task, even a well-learned prompt may not be suitable for all the data instances from a large population~\cite{ScaoR21}, and hence it is non-trivial to design effective transferring strategy considering both task- and instance-level characteristics. 

To address the above issues, we propose \textbf{\textsc{Ptg}}: \textbf{P}rompt Transfer for \textbf{T}ext \textbf{G}eneration, a novel prompt-based transfer learning approach for text generation. \textsc{Ptg} is built upon a transfer learning setting. Specifically, we learn \emph{source prompts} from a number of representative source generation tasks and then transfer these prompts as \emph{target prompts} to perform target generation tasks. The core idea is that these learned source prompts serve as representation bases (\ie \emph{value vectors} in self-attention mechanism). For each data instance from a new task, we learn a specific target prompt by attending to highly relevant source prompts. To support such an approach, we construct a multi-key memory network storing both source prompts and prompt clusters for key-value prompt finding, and then design an adaptive attention mechanism considering both task- and instance-level information to derive the target prompt. Instead of using a fixed prompt for a new task, our approach is able to effectively learn the most suitable prompt representation from source prompts for a specific data instance. Such an adaptive mechanism considers the specific instance-level features, making our approach more flexible to transfer to new text generation tasks.

To the best of our knowledge, we are the first to introduce the idea of prompting in transfer learning to address text generation tasks. For evaluation, we test \textsc{Ptg} on 14 datasets from three sets of text generation tasks: i) \textit{compression} to express salient information in concise text such as summarization; ii) \textit{transduction} to transform text while preserving content precisely such as style transfer; and iii) \textit{creation} to produce new content from input context such as story generation. In both fully-supervised and few-shot experiments, \textsc{Ptg} yields competitive or better results than fine-tuning PLMs. 

Besides performance benefits, more importantly, we release our source prompts to serve as an open-source prompt library. Researchers can train new task prompts added to our library and reuse these learned prompts to improve unseen text generation tasks. Our library can further act as an analysis tool, such as 
analyzing what factors influence prompts' transferability across generation tasks and interpreting the task similarity by measuring the corresponding prompt similarity.

\section{Related Work}

\paratitle{Prompt-based Language Models}. Prompt-based learning is a way of leveraging PLMs by prepending task-specific instructions to the task input when feeding into PLMs. Early approaches mainly utilized hand-crafted prompts to adapt to different generation tasks~\cite{gpt3,RaffelSRLNMZLL20,zou2021controllable}. However, manually designed prompts are not flexible and cannot be applied to more kinds of new tasks. Thus, recent works have focused on automating the learning of discrete prompts~\cite{autoprompt,gao2020making}. 
However, learning prompts over discrete space is hard to optimize and likely to be sub-optimal. To address these problems, many works proposed to optimize continuous prompts~\cite{ptuning,prefix}, which are more flexible to many kinds of tasks. Among these studies, prefix-tuning~\cite{prefix} prepended a sequence of vectors to the input for text generation tasks. By contrast, we utilize soft prompts to investigate transfer learning for text generation and demonstrate that generation tasks can often help each other via prompt transfer.

\paratitle{Transferability of Natural Language Processing}. We are also closely related to existing works on transfer learning in NLP tasks~\cite{JeongSKKYYK20,WieseWN17,Liu0BPS19}. Prior studies have shown that cross-task transfer can address the data scarcity issue~\cite{WieseWN17}, enhance the ability to complex reasoning and inference~\cite{JeongSKKYYK20}, or learn effective word representations~\cite{Liu0BPS19}. 
Efforts to transfer prompts for addressing NLU tasks have also been developed~\cite{spot,tpt}. As a representative work, \citet{spot} used the learned prompt to directly initialize the prompt for a target task while not considering the specific input. Besides, \citet{wang2021learning} uses prompt-based learning in continual learning to sequentially address one-by-one image classification tasks. Our work focuses on prompt transfer between text generation tasks by utilizing prompts to extract implicit task-related knowledge and considering specific model inputs for the most helpful knowledge transfer.
\section{Preliminary}
\label{sec-pre}

\subsection{Problem Formulation}
\label{sec-prompt}

Generally, the objective of text generation is to model the conditional probability $\text{Pr}(y|x)$, where $x=\langle w_1,\dots,w_n \rangle$ and $y=\langle z_1,\dots,z_m \rangle$ denote the input text and output text respectively and consist of sequences of tokens from a vocabulary $\mathcal{V}$. 

Prompting is a technique for injecting extra task information to PLMs as a condition during the generation of output text~\cite{gpt3}. Typically, prompting is conducted by prepending a series of  tokens (discrete prompts) or continuous vectors (continuous prompts) to the input $x$. In our paper, we adopt continuous prompts. Specifically, given a series of $n$ input tokens, $x=\langle w_1,\dots,w_n \rangle$, we first utilize PLM to embed the tokens, forming a matrix $\bm{E}_x \in \mathbb{R}^{n \times e}$, where $e$ is the dimension of the embedding space. Then, our continuous prompt $p$ is represented as a parameter matrix $\bm{E}_p \in \mathbb{R}^{l \times e}$, where $l$ is the number of prompt vectors. The prompt $p$ is then prepended to the embedded input forming a single matrix $[\bm{E}_p;\bm{E}_x] \in \mathbb{R}^{(l+n) \times e}$ which is encoded by PLMs as an ordinary sequence, such that the model maximizes the likelihood of the ground-truth $y$, \ie $\text{Pr}(y|[p;x])$.

\subsection{Prompt-based Transfer Learning}

In a general transfer learning framework, we define a set of source generation tasks $\mathcal{S}=\{\mathcal{S}_1, \dots , \mathcal{S}_T\}$, where the $t$-th task $\mathcal{S}_t=\{(x^t_i, y^t_i)\}^{n_t}_{i=1}$ contains $n_t$ tuples of the input text $x^t_i \in \mathcal{X}$ and its corresponding output text $y^t_i \in \mathcal{Y}$. For a target generation task $\mathcal{T}$, the goal of transfer learning is to use the previously learned task-specific knowledge of the source tasks $\mathcal{S}$ to help improve the performance of a learned model $f_\theta$ (parameterized by $\theta$) in the target task $\mathcal{T}$.

In this paper, we consider a new transfer learning setting based on prompting. Specifically, the parameters of the underlying PLM are frozen, and the text generation tasks have to be fulfilled by prepending prompts (continuous vectors) to input as described in Section~\ref{sec-prompt}. Formally,  we will learn an independent \textit{source prompt} $p_t$ for each source generation task $\mathcal{S}_t$ based on a shared frozen PLM by maximizing the likelihood $\text{Pr}(y^t_i|[p_t;x^t_i])$. Our core idea is to transfer these learned source prompts to a new (target) text generation task, such that the target generation task can be performed in zero or few shot settings. 


\section{Approach}

\begin{figure}[tb]
	\centering
	\includegraphics[width=0.47\textwidth]{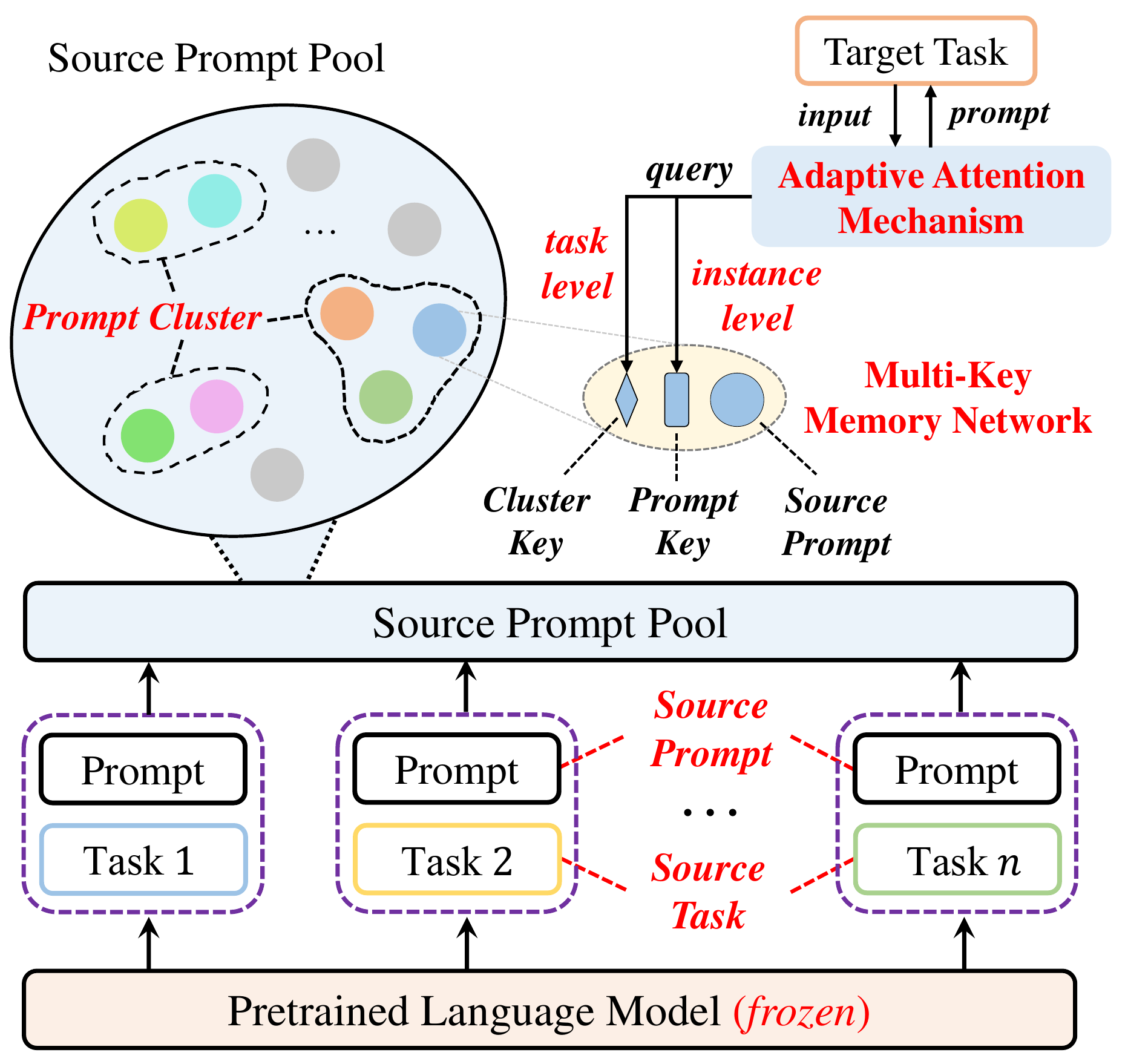}
	\caption{Overview of our proposed model \textsc{Ptg}.}
	\label{fig-model}
	\vspace{-0.3cm}
\end{figure}

Our proposed method, \underline{P}rompt Transfer for \underline{T}ext \underline{G}eneration (\textsc{Ptg}), is depicted in Figure~\ref{fig-model}. Our approach first learns a number of source prompts for various representative source generation tasks, and then derive the prompt for the target generation task with a novel adaptive attention mechanism. Next we will describe each part in detail.


\subsection{Learning Transferable Source Prompts}
\label{sec-prompt-learning}

To extract task-related knowledge from source generation tasks, we learn a set of source prompts and store them in a \textit{source prompt pool}~\cite{wang2021learning}. The motivations for introducing the prompt pool are twofold. First, we expect to identify the similarity between source generation tasks. Second, the pool stores task-specific prompts for every source task, which can be shared by all target tasks. 

\paratitle{Constructing Source Prompt Pool}. For each source generation task $\mathcal{S}_t$, we aim to learn a source prompt $p_t$ given its training data $\{(x^t_i, y^t_i)\}^{n_t}_{i=1}$. Following the learning steps in Section~\ref{sec-prompt}, we learn an independent source prompt $p_t$ for each source task $\mathcal{S}_t$ based on a shared frozen PLM, \ie BART. These source prompts are stored in a prompt pool $\mathcal{P} = \{p_1, \dots, p_t, \dots, p_T\}$, where $T$ is the total number of source text generation tasks.

To construct the source prompt pool, a key point lies in the selection of source text generation tasks. According to the literature~\cite{DengTLXH21}, text generation tasks can be categorized as performing compression, transduction, or creation based on changes in conveyed information from input to output. Moreover, recent studies have shown that few but diverse source tasks/domains also lead to remarkable transfer learning performance~\cite{FriedmanDC21,ZhuangQDXZZXH21}. Therefore, we select six text generation tasks (including 14 public datasets) within the three types of generation tasks for learning their corresponding source prompts.


\paratitle{Clustering Source Prompts}. As described above, the source tasks are diverse in the prompt pool. It is challenging for PLMs to effectively transfer or reuse existing prompts for new tasks. Thus, to identify the similarity between source tasks (prompts), we construct a source prompt pool for more effective cross-task knowledge transfer. In particular, via spectral clustering algorithm~\cite{DingHZGS01}, we group these source prompts into several prompt clusters. Under this algorithm, each prompt $p_t$ is regarded as a node in a weighted undirected graph $\mathcal{G}$. The similarity degree (weight) between node (prompt) $p_i$ and $p_j$ is computed via the position-agnostic Euclidean distances~\cite{tpt}:
\begin{equation}\label{eq-cos}
	w_{i,j} = \frac{1}{1 + \frac{1}{l^2} \sum_{k_1=1}^{l} \sum_{k_2=1}^{l} ||p_{i,k_1} - p_{j,k_2}||},
\end{equation}
where $p_{i,k_1}, p_{j,k_2}$ denote the $k_1$-th and $k_2$-th vector of prompt $p_i$ and $p_j$, respectively. We then adopt the min-max cut strategy~\cite{DingHZGS01} to partition the graph $\mathcal{G}$ into several subgraphs representing different prompt clusters $\mathcal{C} = \{\mathcal{C}_1, \dots, \mathcal{C}_m\}$, where $m$ is the total number of clusters. When transferring the source prompts, it will be better to  identify the suitable prompt cluster and select the most relevant source prompt. By contrast, previous works considered each source prompt equally and ignore the differences between different tasks~\cite{spot,tpt}.

\paratitle{Multi-Key Memory Network}. With source prompts encoding task-related knowledge, the second motivation is to share them with every target generation task. To facilitate the prompt transfer from source tasks to target tasks, we build a multi-key memory network to store these clustered prompts. Specifically, for a source prompt $p_t$ from the prompt cluster $\mathcal{C}_z$, \ie $p_t \in \mathcal{C}_z$, it is associated with a learnable cluster key $\bm{k}^c_z$ and a learnable prompt key $\bm{k}^p_t$, as follows:
\begin{equation}
	\mathcal{\tilde{P}} = \{\mathcal{C}_z: \langle \bm{k}^c_z, \bm{k}^p_t, p_t \rangle\}_{z=1}^m,
\end{equation}
where $\bm{k}^c_z, \bm{k}^p_t \in \mathbb{R}^d$, and $d$ is the key embedding size. In our memory network, these learned source prompts serve as representation bases, \ie value vectors, which can be transferred to target generation tasks through key-value prompt finding.


\subsection{Transferring Instance Adaptive Prompts}
\label{sec-prompt-transfer}

Previous works~\cite{prefix,spot} usually consider only the task information but ignore the specific input data when deriving prompts. However, for a single task, even a well-learned prompt may not be suitable for all the data instances~\cite{ScaoR21}, and thus it is non-trivial to design effective transferring strategy considering both task- and instance-level characteristics. In our model, we design an adaptive attention mechanism to incorporate the instance feature for constructing the target prompt.

\paratitle{Adaptive Attention Mechanism}. Specifically, for an instance $(x, y)$ of the target task $\mathcal{T}$, we use both task-level and instance-level queries to adaptively lookup and select the source prompts for transferring the previously learned task-related knowledge. The task-level query aims to select the overall information related to the specific target task, which is defined as a learnable task query vector $\bm{q}^{task} \in \mathbb{R}^d$. However, the source prompts in the pool are diverse but limited, thus the task-level prompt may not well adapt to all the data instances of the target generation task. Therefore, we design an instance-level query to learn the target prompt by attending to the highly relevant source prompts to help improve the model performance in specific instances. The instance-level query is computed as the input encoding $\bm{q}^{ins} \in \mathbb{R}^d$ through a frozen PLM such as BERT~\cite{DevlinCLT19}: 
\begin{equation}\label{eq-bert}
	\bm{q}^{ins}=\text{Average}(\text{BERT}(x)),
\end{equation}
where we average the top-layer representations of every input tokens encoded by BERT. 

For a source prompt $p_t \in \mathcal{C}_z$, we use $\bm{q}^{task}$ and $\bm{q}^{ins}$ to lookup its corresponding cluster key and source key respectively, following multi-head attention~\cite{VaswaniSPUJGKP17}. Thus, the final matching score between the instance $x$ and prompt $p_t$ is calculated as:
\begin{equation}\label{mha}
	s_t = \text{softmax}(\lambda \cdot \bm{q}^{task\top} \bm{k}^{c}_z + (1-\lambda)\cdot \bm{q}^{ins\top} \bm{k}^{p}_t), 
\end{equation}
where $\lambda$ is a hyper-parameter. Finally, according to the weight score, the selected source prompt is computed as: $\tilde{p} = \sum_{t=1}^{T} s_t \cdot p_t$.

Compared to other prompt-based transfer learning methods that used only a fixed prompt for a new task~\cite{spot,prefix}, our adaptive attention mechanism is able to effectively learn the most suitable prompt representation from source prompts for a specific data instance. Such a mechanism makes our model more flexible to transfer to new text generation tasks.

\paratitle{Prompt-based Text Generation}. Based on the above adaptive attention mechanism, we retrieve the prompt $\tilde{p}$ encoding the most useful and relevant knowledge to help the model perform the specific generation instances. As described in Section~\ref{sec-prompt}, we prepend the prompt $\tilde{p}$ to the input embedding of $x$, which then flows through a generative PLM such as BART~\cite{LewisLGGMLSZ20} for generating text. The generative PLM is optimized via maximum likelihood estimation (MLE) as:
\begin{equation}
	\mathcal{L}_{\text{MLE}}(\theta) = \mathbb{E}_{(x,y) \sim (\mathcal{X}, \mathcal{Y})} \log\text{Pr}(y|[\tilde{p};x]).
\end{equation}

During the learning process of the target task, the retrieved prompt $\tilde{p}$ is adaptive to different instances and is frozen because it encodes the previously \textit{learned} task-related knowledge.

\subsection{Model Discussion}


For prompt-based transfer learning in text generation, the key point lies in how to effectively transfer or reuse existing prompts (encoding task-specific knowledge) for new generation tasks considering both task- and instance-level characteristics.

To achieve this goal, we first learn a set of source prompts encoding task-specific knowledge from a number of representative source text generation tasks (Section~\ref{sec-prompt-learning}). These source prompts serve as representation bases, \ie value vectors in the multi-key memory network. Moreover, we design an adaptive attention mechanism considering both task- and instance-level information for constructing the target prompt (Section~\ref{sec-prompt-transfer}). Each data instance from a new generation task can learn a specific prompt by attending to the most highly relevant source prompts.

Compared with typical transfer learning methods, our model utilizes a lightweight technique, \ie prompting, to learn task-specific knowledge from source tasks. Our pretrained source prompts can help PLMs perform more effective and useful knowledge transfer.



\section{Experiments}

In this section, we first set up the experiments, and then report the
results and analysis.


\subsection{Experimental Setup}
\paratitle{Datasets}. We select 14 public datasets divided into three types of text generation tasks: i) \textit{compression} to express salient information in concise text including summarization (CNN/Daily Mail~\cite{SeeLM17}, XSum~\cite{NarayanCL18}, MSNews~\cite{LiuYGQZJCFSGWCJ21}, Multi-News~\cite{FabbriLSLR19}, NEWSROOM~\cite{GruskyNA18}) and question generation (SQuAD~\cite{RajpurkarZLL16}); ii) \textit{transduction} to transform text while preserving content precisely including style transfer (Wiki Neutrality~\cite{PantDM20}) and text paraphrase (Quora~\cite{WangHF17}); and iii) \textit{creation} to produce new content from input context including dialog (PersonaChat~\cite{KielaWZDUS18}, TopicalChat~\cite{Gopalakrishnan2019}, DailyDialog~\cite{LiSSLCN17}, DSTC7-AVSD~\cite{alamri2019audiovisual}, MultiWOZ~\cite{budzianowski2018large}) and story generation (WritingPrompts~\cite{LewisDF18}). Dataset statistics are in Appendix~\ref{app-dataset}. 

\paratitle{Baselines}. We compare our proposed \textsc{Ptg} to the following baselines:

\textbullet~\textbf{\textsc{Gpt-2}}~\cite{radford2019language}, \textbf{\textsc{Bart}}~\cite{LewisLGGMLSZ20}, and \textbf{\textsc{T5}}~\cite{RaffelSRLNMZLL20}: These are three representative PLMs for text generation, where all pretrained parameters are fine-tuned on each target task dataset separately. We adopt the \textsc{Large} version of these PLMs.

\textbullet~\textbf{\textsc{PrefixTuning}}~\cite{prefix}: It is the recent state-of-the-art prompt-based PLM for text generation by concatenating a sequence of vectors and the input, which keeps PLM parameters frozen but optimizes a set of continuous prefix vectors.

\textbullet~\textbf{\textsc{SPoT}}~\cite{spot}: It also adopts a prompt-based transfer learning method which first trains a prompt on source tasks and then uses the resulting prompt to initialize the prompt for a target task.

\textbullet~\textbf{\textsc{Multi-task ModelTuning}}: This strong multi-task baseline first fine-tunes \textbf{\textsc{Bart}}
on the same source tasks used for \textsc{Ptg} and then fine-tunes on each target task dataset individually.

\begin{table*}[t]
\renewcommand\arraystretch{1.1}
\begin{minipage}[t]{\linewidth}
  \centering
     \makeatletter\def\@captype{table}\makeatother\small
     \begin{tabular}{l r r r r r r r}
			\toprule[1pt]
			\textbf{Target Task} & \multicolumn{3}{c}{\textsc{Summarization} (CNN/Daily Mail)} & \multicolumn{4}{c}{\textsc{Dialog} (PersonaChat)} \\
			\cmidrule(r){1-1}\cmidrule(r){2-4}\cmidrule(r){5-8}
			\textbf{\#Metrics} & ROUGE-1 & ROUGE-2 & ROUGE-L & BLEU-1 & BLEU-2 & Distinct-1 & Distinct-2 \\
			\midrule[0.5pt]
			\textsc{\textbf{Gpt-2}}$_\textsc{Large}$ & 30.30 & 7.66 & 26.40 & 36.07 & 22.64 & \textbf{1.57} & \textbf{8.54} \\
			\textsc{\textbf{Bart}}$_\textsc{Large}$ & 41.37 & 21.16 & 38.36 & 40.48 & 26.48 & 1.42 & 7.60  \\
			\textsc{\textbf{T5}}$_\textsc{Large}$ & 40.47 & 20.30 & 37.57 & \underline{42.23} & \underline{27.36} & 1.39 & 7.63  \\
			\textsc{\textbf{PrefixTuning}} & \underline{41.79} & 20.69 & \underline{38.50} & 41.87 & 27.28 & 1.33 & 7.20\\
			\textbf{\textsc{SPoT}} & 39.38 & 17.24 & 36.71 & 39.74 & 26.52 & 1.33 & 7.81 \\
			\textsc{\textbf{MT ModelTuning}} & 41.43 & \underline{21.17} & 38.40 & 40.47 & 26.49 & 1.45 & 7.83 \\
			\cline{1-8}
			\textsc{\textbf{Ptg}} & \textbf{42.40} & \textbf{21.35} & \textbf{39.14} & \textbf{45.46} & \textbf{29.52} & \underline{1.46} & \underline{8.34} \\
			\bottomrule[1pt]
		\end{tabular}
		\caption{Cross-task transferability performance comparisons of different methods in fully-supervised setting. \textbf{Bold} and \underline{underline} fonts denote the best and the second best methods (the same as below).\\}
		\label{tab:full-task}
  \end{minipage}
  \begin{minipage}[t]{\linewidth}
   \centering
        \makeatletter\def\@captype{table}\makeatother\small
        \begin{tabular}{l r r r r r r r}
			\toprule[1pt]
			\textbf{Target Dataset} & \multicolumn{3}{c}{\textsc{CNN/Daily Mail}} & \multicolumn{4}{c}{\textsc{PersonaChat}} \\
			\cmidrule(r){1-1}\cmidrule(r){2-4}\cmidrule(r){5-8}
			\textbf{\#Metrics} & ROUGE-1 & ROUGE-2 & ROUGE-L & BLEU-1 & BLEU-2 & Distinct-1 & Distinct-2 \\
			\midrule[0.5pt]
			\textsc{\textbf{Gpt-2}}$_\textsc{Large}$ & 30.30 & 7.66 & 26.40 & 36.07 & 22.64 & \textbf{1.57} & \textbf{8.54} \\
			\textsc{\textbf{Bart}}$_\textsc{Large}$ & 41.37 & 21.16 & 38.36 & 40.48 & 26.48 & 1.42 & 7.60  \\
			\textsc{\textbf{T5}}$_\textsc{Large}$ & 40.47 & 20.30 & 37.57 & 42.23 & 27.36 & 1.39 & 7.63  \\
			\textsc{\textbf{PrefixTuning}} & \underline{41.79} & 20.69 & 38.50 & 41.87 & 27.28 & 1.33 & 7.20\\
			\textbf{\textsc{SPoT}} & 39.85 & 18.21 & 36.33 & 40.39 & 26.34 & 1.32 & 7.60 \\
			\textsc{\textbf{MT ModelTuning}} & 41.71 & \underline{21.41} & \underline{38.67} & \underline{42.53} & \underline{27.83} & 1.39 & 7.86 \\
			\cline{1-8}
			\textsc{\textbf{Ptg}} & \textbf{42.68} & \textbf{21.63} & \textbf{39.45} & \textbf{45.47} & \textbf{29.52} & \underline{1.43} & \underline{8.34} \\
			\toprule[1pt]
			\textbf{Target Dataset} & \multicolumn{3}{c}{\textsc{XSum}} & \multicolumn{4}{c}{\textsc{DailyDialog}} \\
			\cmidrule(r){1-1}\cmidrule(r){2-4}\cmidrule(r){5-8}
			\textbf{\#Metrics} & ROUGE-1 & ROUGE-2 & ROUGE-L & BLEU-1 & BLEU-2 & Distinct-1 & Distinct-2 \\
			\midrule[0.5pt]
			\textsc{\textbf{Gpt-2}}$_\textsc{Large}$ & 28.28 & 9.17 & 22.29 & 29.14 & 18.01 & \textbf{5.78} & 21.52 \\
			\textsc{\textbf{Bart}}$_\textsc{Large}$ & \underline{43.93} & \underline{20.78} & \underline{35.94} & 32.62 & 21.77 & 5.16 & 25.08  \\
			\textsc{\textbf{T5}}$_\textsc{Large}$ & 41.01 & 17.84 & 32.60 & 31.54 & 20.08 & \underline{5.70} & \underline{29.25}  \\
			\textsc{\textbf{PrefixTuning}} & 42.87 & 19.98 & 34.82 & 34.00 & 21.63 & 4.31 & 19.95\\
			\textbf{\textsc{SPoT}} & 41.43 & 17.56 & 31.33 & 30.22 & 20.11 & 4.91 & 25.56 \\
			\textsc{\textbf{MT ModelTuning}} & 43.75 & 20.70 & 35.66 & \underline{34.41} & \underline{23.08} & 5.46 & 27.23 \\
			\cline{1-8}
			\textsc{\textbf{Ptg}} & \textbf{44.21} & \textbf{20.99} & \textbf{36.00} & \textbf{42.72} & \textbf{28.75} & 5.36 & \textbf{29.48} \\
			\bottomrule[1pt]
		\end{tabular}
		\caption{Cross-dataset transferability performance comparisons of different methods in fully-supervised setting.\\}
		\label{tab:full-dataset}
   \end{minipage}
\end{table*}

We conduct all methods in the same setting to obtain their results without special tricks such as label smoothing. Compared with other baselines, our model is extremely lightweight, \ie when solving target generation tasks, we freeze the transferred target prompt and parameters of the backbone PLM but only tune the multi-head attention parameters in adaptive attention mechanism (Eq.~\ref{mha}). 

In particular, we adopt BART-\textsc{large} to learn a set of source prompts. The length of prompt is set to $200$ and the learning rate is set to $1 \times 10^{-3}$. For the target generation task, we utilize BART-\textsc{large} as the generation backbone and frozen BERT-\textsc{large} to obtain the instance-level query $\bm{q}^{ins}$. The dimension $d$ is set to $1024$, which is the same as the embedding size $e$ of the BERT/BART-\textsc{large}. The multi-head attention in adaptive attention mechanism has $16$ heads. During fine-tuning, the learning rate of BART is set to $3 \times 10^{-5}$ and the learning rate of cluster key $\bm{k}^{c}$, prompt key $\bm{k}^{p}$, task key $\bm{q}^{task}$ and multi-head attention is set to $1 \times 10^{-3}$. The value of $\lambda$ is set to 0.5 based on the performance in validation set. The training details of baselines can be found in Appendix~\ref{app-configuration}.


\paratitle{Evaluation Metrics}. For performance comparison, we adopt three automatic evaluation metrics widely used by previous works, \ie BLEU~\cite{papineni2002bleu},  ROUGE~\cite{lin2004rouge} and Distinct~\cite{LiGBGD16}. Specifically, BLEU-$n$ measures the ratios of the co-occurrences of $n$-grams between the generated and real text; ROUGE-$n$ measures the text quality by counting the overlapping $n$-grams between the generated and real text; and Distinct-$n$ measures the degree of diversity by calculating the number of distinct $n$-grams in generated text.

\begin{table*}[t]
\renewcommand\arraystretch{1.1}
\setlength\tabcolsep{3pt}
\begin{minipage}[t]{\linewidth}
  \centering
     \makeatletter\def\@captype{table}\makeatother\scriptsize
     \begin{tabular}{l r r r r r r r r}
			\toprule[1pt]
			\textbf{Target Task} & \multicolumn{4}{c}{\textsc{Summarization} (R-1/R-2/R-L)} & \multicolumn{4}{c}{\textsc{Dialog} (B-1/B-2/D-1/D-2)} \\
			\cmidrule(r){1-1}\cmidrule(r){2-5}\cmidrule(r){6-9}
			\textbf{\#Instances} & \multicolumn{1}{c}{50} & \multicolumn{1}{c}{100} & \multicolumn{1}{c}{200} & \multicolumn{1}{c}{500} & \multicolumn{1}{c}{50} & \multicolumn{1}{c}{100} & \multicolumn{1}{c}{200} & \multicolumn{1}{c}{500} \\
			\midrule[0.5pt]
			\textsc{\textbf{Gpt-2}}$_\textsc{Large}$ & 19.8/\;\,3.1/17.5 & 20.6/\;\,3.9/18.4 & 26.1/\;\,6.1/23.3 & 29.2/\;\,7.3/26.0 & 27.7/10.4/2.2/10.9 & 27.6/10.4/2.2/11.4 & 29.3/11.1/2.1/11.5 & 31.4/12.0/1.9/10.7 \\
			\textsc{\textbf{Bart}}$_\textsc{Large}$ & 37.5/16.9/34.4 & 38.8/17.9/35.6 & 39.3/18.4/36.1 & 39.9/19.0/36.7 & 22.7/\;\,9.0/1.3/\;\,5.4 & 30.0/11.9/1.3/\;\,5.2 & 32.4/12.8/1.3/\;\,5.7 & 31.7/12.6/1.3/\;\,5.6  \\
			\textsc{\textbf{T5}}$_\textsc{Large}$ & 39.1/18.3/36.2 & 39.9/18.5/36.8 & 40.0/18.7/37.0 & 39.6/19.2/36.7 & 41.7/15.5/0.9/\;\,6.6 & 42.1/15.7/0.8/\;\,5.4 & 43.1/16.3/0.7/\;\,4.6 & 45.1/17.4/0.8/\;\,4.4  \\
			\textsc{\textbf{PrefixT}} & 32.2/12.4/28.5 & 32.3/12.5/28.5 & 34.0/13.7/30.9 & 37.5/16.3/34.7 & 39.6/23.9/0.6/\;\,3.4 & 39.7/24.0/0.5/\;\,3.1 & 36.4/22.4/0.8/\;\,3.7 & 25.7/16.1/1.1/\;\,4.1 \\
			\textbf{\textsc{SPoT}} & 31.3/11.8/27.5 & 31.9/11.8/27.5 & 33.6/12.6/29.3 & 36.5/16.0/33.6 &
			38.3/22.1/0.5/\;\,3.0 & 38.2/22.0/0.5/\;\,3.0 &
			39.0/23.2/0.8/\;\,4.1 & 41.1/23.5/1.0/\;\,4.5 \\
			\textsc{\textbf{ModelT}} & 36.2/15.6/32.8 & 37.8/16.6/34.4 & 38.6/17.3/35.2 & 39.3/17.9/35.8 & 24.9/\;\,9.9/1.5/\;\,6.6 & 24.8/\;\,9.8/1.6/\;\,6.6 & 27.8/11.0/1.6/\;\,7.1 & 28.9/11.4/1.7/\;\,7.8 \\
			\cline{1-9}
			\textsc{\textbf{Ptg}} & \underline{37.8}/16.7/\underline{34.5} & \underline{39.0}/17.5/\underline{35.6} & \underline{39.3}/17.7/\underline{36.2} & \textbf{40.1}/\underline{19.1}/\textbf{36.8} & 37.3/\underline{22.6}/1.1/\;\,6.2 & \underline{39.9}/\underline{21.2}/1.1/\;\,5.3 & \underline{37.7}/\textbf{23.6}/1.1/\;\,4.9 & \underline{37.7}/\textbf{24.2}/1.4/\;\,6.3 \\
			\bottomrule[1pt]
		\end{tabular}
		\caption{Cross-task transferability performance comparisons of different methods in few-shot setting. B-$n$, R-$n$, D-$n$, and \textsc{ModelT} are short for BLEU, ROUGE, Distinct and \textsc{Multi-task ModelTuning} (the same as below). \\}
		\label{tab:fs-task}
  \end{minipage}
  \begin{minipage}[t]{\linewidth}
   \centering
        \makeatletter\def\@captype{table}\makeatother\scriptsize
        \begin{tabular}{l r r r r r r r r}
			\toprule[1pt]
			\textbf{Target Data} & \multicolumn{4}{c}{\textsc{CNN/Daily Mail} (R-1/R-2/R-L)} & \multicolumn{4}{c}{\textsc{PersonaChat} (B-1/B-2/D-1/D-2)} \\
			\cmidrule(r){1-1}\cmidrule(r){2-5}\cmidrule(r){6-9}
			\textbf{\#Instances} & \multicolumn{1}{c}{50} & \multicolumn{1}{c}{100} & \multicolumn{1}{c}{200} & \multicolumn{1}{c}{500} & \multicolumn{1}{c}{50} & \multicolumn{1}{c}{100} & \multicolumn{1}{c}{200} & \multicolumn{1}{c}{500} \\
			\midrule[0.5pt]
			\textsc{\textbf{Gpt-2}}$_\textsc{Large}$ & 19.8/\;\,3.1/17.5 & 20.6/\;\,3.9/18.4 & 26.1/\;\,6.1/23.3 & 29.2/\;\,7.3/26.0 & 27.7/10.4/2.2/10.9 & 27.6/10.4/2.2/11.4 & 29.3/11.1/2.1/11.5 & 31.4/12.0/1.9/10.7 \\
			\textsc{\textbf{Bart}}$_\textsc{Large}$ & 37.5/16.9/34.4 & 38.8/17.9/35.6 & 39.3/18.1/36.1 & 39.9/19.0/36.7 & 22.7/\;\,9.0/1.3/\;\,5.4 & 30.0/11.9/1.3/\;\,5.2 & 32.4/12.8/1.3/\;\,5.7 & 31.7/12.6/1.3/\;\,5.6  \\
			\textsc{\textbf{T5}}$_\textsc{Large}$ & 39.1/18.3/36.2 & 37.9/18.5/36.8 & 39.0/18.7/36.0 & 39.6/19.2/36.7 & 31.7/15.5/0.9/\;\,6.6 & 32.1/15.7/0.8/\;\,5.4 & 33.1/16.3/0.7/\;\,4.6 & 35.1/17.4/0.8/\;\,4.4  \\
			\textsc{\textbf{PrefixT}} & 32.2/12.4/28.5 & 32.3/12.5/28.5 & 34.0/13.7/30.9 & 37.5/16.3/34.7 & 39.6/23.9/0.6/\;\,3.4 & 39.7/24.0/0.5/\;\,3.1 & 36.4/22.4/0.8/\;\,3.7 & 25.7/16.1/1.1/\;\,4.1 \\
			\textbf{\textsc{SPoT}} & 31.9/11.5/26.8 & 31.9/11.4/26.8 & 33.0/12.8/29.3 & 36.6/15.5/33.2 &
			37.6/22.0/0.5/\;\,3.1 & 37.6/22.2/0.5/\;\,3.2 &
			35.0/20.2/0.7/\;\,3.2 & 21.2/15.6/1.0/\;\,3.8 \\
			\textsc{\textbf{ModelT}} & 37.7/17.0/34.5 & 38.8/17.9/35.6 & 39.3/18.2/36.0 & 40.5/19.0/36.1 & 32.0/13.1/2.4/12.4 & 34.2/13.9/2.2/11.9 & 35.9/14.7/2.1/11.7 & 35.5/14.7/2.0/10.8 \\
			\cline{1-9}
			\textsc{\textbf{Ptg}} & \underline{37.9}/16.5/\underline{34.5} & \underline{38.7}/17.5/\underline{35.8} & \textbf{39.5}/\underline{18.3}/\textbf{36.2} & \underline{39.9}/18.7/\underline{36.6} & \underline{34.6}/\underline{21.5}/1.1/\;\,4.5 & \underline{36.9}/\underline{19.3}/1.0/\;\,5.5 & \textbf{38.6}/\textbf{24.1}/1.0/\;\,4.4 & \textbf{36.7}/\textbf{23.0}/1.2/\;\,5.5 \\
			\toprule[1pt]
			\textbf{Target Data} & \multicolumn{4}{c}{\textsc{XSum} (R-1/R-2/R-L)} & \multicolumn{4}{c}{\textsc{DailyDialog} (B-1/B-2/D-1/D-2)} \\
			\cmidrule(r){1-1}\cmidrule(r){2-5}\cmidrule(r){6-9}
			\textbf{\#Instances} & \multicolumn{1}{c}{50} & \multicolumn{1}{c}{100} & \multicolumn{1}{c}{200} & \multicolumn{1}{c}{500} & \multicolumn{1}{c}{50} & \multicolumn{1}{c}{100} & \multicolumn{1}{c}{200} & \multicolumn{1}{c}{500} \\
			\midrule[0.5pt]
			\textsc{\textbf{Gpt-2}}$_\textsc{Large}$ & 12.2/\;\,1.5/\;\,9.8 & 11.3/\;\,1.1/\;\,9.1 & 11.1/\;\,1.1/\;\,8.9 & 12.9/\;\,1.7/10.2 & 18.5/\;\,7.0/5.9/23.3 &  19.3/\;\,7.3/5.6/22.8 & 20.9/\;\,7.9/5.4/22.0 & 22.0/\;\,8.3/5.5/\;\,2.9 \\
			\textsc{\textbf{Bart}}$_\textsc{Large}$ & 33.2/10.3/25.2 & 32.8/11.0/26.6 & 34.5/11.6/25.5 & 36.4/13.2/28.2 & 22.0/\;\,8.5/3.5/15.6 & 22.2/\;\,8.5/3.3/14.5 & 24.8/\;\,9.6/3.4/14.9 & 24.3/\;\,9.4/3.8/11.4  \\
			\textsc{\textbf{T5}}$_\textsc{Large}$ & 23.2/\;\,5.0/16.6 & 23.4/\;\,5.3/17.1 & 26.0/\;\,7.1/19.5 & 30.8/10.3/24.2 & 30.6/14.8/2.5/14.9 & 41.0/15.0/2.4/14.1 & 30.9/15.1/2.8/15.4 & 30.6/15.1/3.2/17.7  \\
			\textsc{\textbf{PrefixT}} & 25.0/\;\,8.3/17.9 & 25.0/\;\,8.2/17.9 & 25.1/\;\,8.2/18.1 & 27.5/\;\,9.8/19.7 & 38.1/22.6/2.8/14.1 & 38.4/22.8/2.5/12.0 & 35.2/21.0/2.5/11.6 & 21.8/13.5/3.8/16.1 \\
			\textbf{\textsc{SPoT}} & 23.4/\;\,6.6/16.6 & 23.4/\;\,6.5/16.6 & 23.5/\;\,6.8/17.0 & 25.5/\;\,7.5/18.6 & 35.5/20.6/2.5/13.2 & 35.7/20.8/2.3/12.8 & 33.6/18.9/2.2/11.9 & 25.0/13.2/3.7/16.1 \\
			\textsc{\textbf{ModelT}} & 35.6/13.1/27.8 & 35.7/13.3/28.0 & 36.0/13.6/28.4 & 36.1/13.8/28.5 & 28.2/11.1/5.4/24.3 & 30.4/11.8/5.2/23.9 & 29.8/11.8/4.9/23.0 & 29.5/11.7/4.7/22.3 \\
			\cline{1-9}
			\textsc{\textbf{Ptg}} & \underline{33.6}/\underline{10.9}/\underline{25.4} & \underline{33.8}/\underline{11.2}/25.9 & \underline{34.7}/\underline{12.0}/\underline{26.8} & \textbf{36.8}/\underline{13.6}/27.7 & \underline{31.8}/\underline{19.4}/2.5/11.6 & 30.9/\underline{18.9}/2.8/12.8 & \underline{31.5}/\underline{19.3}/2.9/13.9 & \textbf{31.0}/\textbf{19.0}/3.1/14.9 \\
			\bottomrule[1pt]
		\end{tabular}
		\caption{Cross-dataset transferability performance comparisons of different methods in few-shot setting.\\}
		\label{tab:fs-dataset}
   \end{minipage}
\end{table*}

\subsection{Fully-Supervised Setting}
\label{sec-full}

Table~\ref{tab:full-task} and Table~\ref{tab:full-dataset} present the fully-supervised  results of cross-task and cross-dataset transferability, respectively, for our model and baselines. In fully-supervised setting, we use all training instances of the target task to train our model. 

For the \textit{cross-task} experiment, we consider two pairs of source and target tasks transfer: 1) the target task is summarization (CNN/Daily Mail), and the source tasks are the mixture of other five tasks; and 2) the target task is dialog (PersonChat), and the source tasks are other five tasks. For the \textit{cross-dataset} experiment, we consider datasets within summarization and dialog. For summarization, the target dataset is CNN/Daily Mail or XSum, and the source datasets are the mixture of other four summarization datasets. For dialog, the target dataset is PersonaChat or DailyDialog, and the source datasets are other four dialog datasets. 

First, by transferring prompts from source tasks to the target task, \textsc{Ptg} outperforms \textsc{Gpt-2}, \textsc{Bart}, \textsc{T5} and \textsc{PrefixTuning}. The results suggest that prompt transfer in \textsc{Ptg} provides an effective means of improving the performance of typical fine-tuning and prompt methods since our method utilizes the knowledge learned from source tasks.

Second, \textsc{Ptg} performs better than the prompt-based transfer method, \textsc{SPoT}. While transferring prompts, \textsc{SPoT} considers each source task equally and ignored the specific instance information. And \textsc{SPoT} only learns a common prompt for source tasks to directly initialize the target prompt. By constrast, \textsc{Ptg} clusters diverse source prompts and uses an adaptive attention mechanism considering both task- and instance-level characteristics. 

Finally, \textsc{Ptg} produces competitive performance or even exceeds the strong \textsc{Multi-task ModelTuning}. Different from most NLU tasks sharing some common knowledge to understand the semantics and syntax of surface words, text generation tasks need to generate diverse text based on different input data, thus having large task boundaries. Thus, in cross-task transfer, simply tuning PLMs on a mixture of tasks without considering the task similarity leads to a performance decrease. While, our prompt-based transfer learning approach can still achieve the best performance, showing that \textsc{Ptg} improves stability across tasks and datasets.

\subsection{Few-Shot Setting} 

In few-shot setting, we only sample a handful of training instances of the target task to train our model. Specificlly,
we subsample the target task dataset to obtain small training datasets of size \{50, 100, 200, 500\}. For each size, we sample 5 different datasets and average over 2 training random seeds. Thus, we average over 10 models for each few-shot setting. In few-shot setting, we adopt the same cross-task and cross-dataset experiments with the fully-supervised setting. Table~\ref{tab:fs-task} and \ref{tab:fs-dataset} shows the few-shot results of our model and baselines.

We can clearly see that \textsc{Ptg} achieves competitive (\underline{underline} fonts) or better performance (\textbf{bold} fonts) than the strong baseline (\ie \textsc{Multi-task ModelTuning}) in most low-data regimes, but the gap narrows as the training dataset size increases. In addition, our model outperforms most of vanilla PLMs in most cases. The reason behind this might be that large PLMs can easily suffer from over-fitting during few-shot learning due to their massive parameters~\cite{gao2020making}. While, in our framework, we adopt a lightweight technique, \ie prompting, to learn source prompts, which can provide the previously learned knowledge in source tasks to PLMs and serve as a better starting point when solving the target tasks.

\begin{table*}[t]
\small
\centering
	\begin{tabular}{l r r r r r r r}
			\toprule[1pt]
			\textbf{Target Task} & \multicolumn{3}{c}{\textsc{Summarization} (CNN/Daily Mail)} & \multicolumn{4}{c}{\textsc{Dialog} (PersonaChat)} \\
			\cmidrule(r){1-1}\cmidrule(r){2-4}\cmidrule(r){5-8}
			\textbf{Model} & ROUGE-1 & ROUGE-2 & ROUGE-L & BLEU-1 & BLEU-2 & Distinct-1 & Distinct-2 \\
			\midrule[0.5pt]
			\textsc{\textbf{Ptg}} w/o Prompt Pool & 41.46 & 20.40 & 38.40 & 39.70 & 24.45 & 0.77 & 4.00 \\
			\textsc{\textbf{Ptg}} w/o Prompt Cluster & 42.10 & 21.15 & 38.86 & 44.63 & 29.20 & 1.34 & 7.78  \\
			\textsc{\textbf{Ptg}} w/o Multi-Key Memory & 42.12 & 21.14 & 38.85 & 44.67 & 29.34 & 1.42 & 8.23\\
			\textsc{\textbf{Ptg}} w/o Instance-level Query & 42.16 & 21.22 & 38.93 & 44.74 & 29.28 & 1.36 & 7.80  \\
			\cline{1-8}
			\textsc{\textbf{Ptg}} & 42.40 & 21.35 & 39.14 & 45.46 & 29.52 & 1.46 & 8.43 \\
			\bottomrule[1pt]
		\end{tabular}
	\caption{Ablation analysis on cross-task transferability experiments.} 
	\label{tab:ablation-results}
\end{table*}

\subsection{Effectiveness of Core Designs}

We further conduct ablation studies to demonstrate the effectiveness of the core designs of \textsc{Ptg}.

\paratitle{Source Prompt Pool}. To confirm the importance of the prompt pool, we design a counterpart of our method with only training a  sharing prompt for all source tasks. From Table~\ref{tab:ablation-results} (row 1), we can see that \textsc{Ptg} significantly outperforms its counterpart with a single prompt, suggesting that the prompt pool encodes task-specific knowledge well.

\paratitle{Source Prompt Cluster}. We remove the step of grouping source prompts into different clusters and directly lookup source prompts based on queries (see in Table~\ref{tab:ablation-results} row 2). The decrease in performance demonstrates that when tasks are diverse, clustering task prompts can identify the similarity between source tasks, thus promoting effective knowledge transfer.

\paratitle{Multi-Key Memory Network}. We remove the learnable key vector associated with prompts and directly transfer the mean of the source prompts to the target task. From Table~\ref{tab:ablation-results} (row 4), we can see this results in a significant drop, demonstrating the importance of introducing learnable keys to dynamically select prompts through query-key matching.

\paratitle{Instance-level Query}. The instance-level query is used in adaptive attention mechanism. When we remove it (Table~\ref{tab:ablation-results} row 3), we only use the task-level query to select source prompts. The declined performance demonstrates that incorporating the instance-level features can indeed help to transfer the most helpful knowledge to the specific instances in target tasks.

\begin{figure}[tb]
	\centering
	\includegraphics[width=0.43\textwidth]{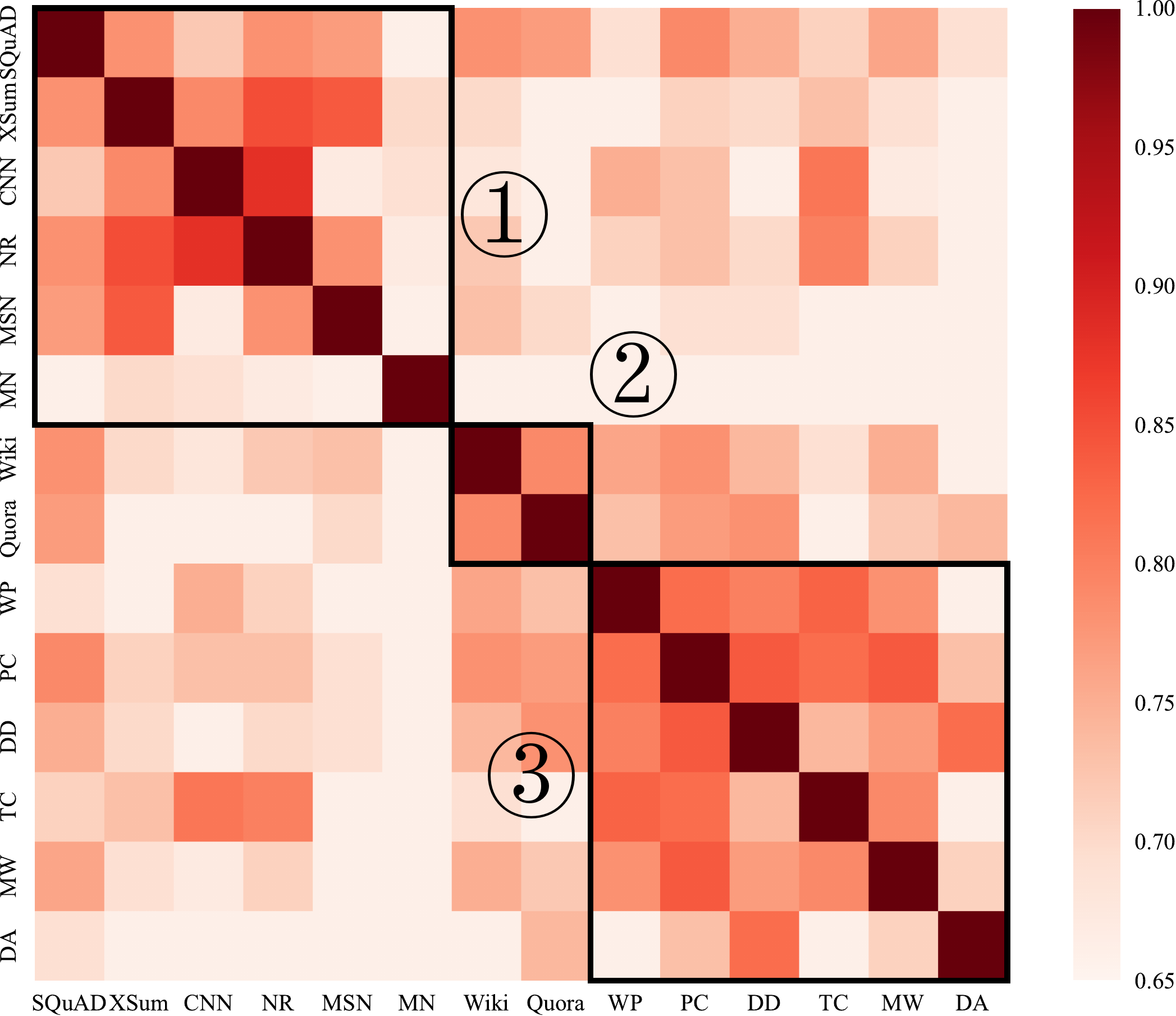}
	\caption{Similarity analysis of 14 datasets within our six generation tasks.}
	\label{fig-similarity}
\end{figure}

\subsection{Task Similarity Analysis}

Figure~\ref{fig-similarity} shows a clustered heatmap of cosine similarities between the source prompts of the 14 public datasets within our six text generation tasks using the position-agnostic Euclidean distances defined by Eq.~\ref{eq-cos}. We can clearly observe that our learned 14 source prompts are roughly grouped into three clusters. Similar tasks and datasets are grouped together into clusters in this heatmap, and these clusters capture many intuitive task relationships. Specifically, these three clusters mainly focus on \textit{compression}, \textit{transduction}, and \textit{creation} tasks respectively. For example, story generation (WritingPrompts) and dialog (PersonaChat) are grouped together into the third cluster. This observation further verifies our conclusion that text generation tasks can help each other within our approach by learning task-specific prompts and then transferring them to the target task. The results also suggest that our method can serve as an effective means of predicting task transferability.

%
%

\section{Conclusion}

This paper presented a prompt-based transfer learning approach for text generation. We learn a set of prompts from a number of representative source generation tasks and then transfer these prompts as target prompts to perform the target generation tasks. In our model, we design an adaptive attention mechanism considering both task- and instance-level information to construct the target prompts. Experiments in fully-supervised and few-shot settings demonstrate the effectiveness of our prompt-based transfer learning model. In future work, we will consider incorporating more kinds of text generation tasks.

\section{Ethical Concerns}

Text generation techniques has been applied to a wide range of meaningful applications for society, such as
game narrative generation, news report generation, and weather report generation. However, this technique may be potentially utilized for harmful applications. Our work improves the quality of generated text compared with traditional methods. Thus, the high-quality text generated by our work makes it difficult to distinguish synthetic text from human-written text, such as fake news and stories. Here we are primarily concerned with two potential ethical issues: the possibility of deliberate misuse of our methodology and the issue of bias.

First, it is somewhat challenging to anticipate the harmful usages of our method since they often involve repurposing our model in a totally different setting or for an unexpected purpose than we planned. To alleviate this issue, we can ask for the assistance of classic security risk assessment frameworks such as detecting threats. Second, biases in training data may cause our model to generate stereotyped or prejudiced texts. This is a worry since the model bias has the potential to hurt some persons in relevant groups in unforeseen ways. To avoid prejudice, it may be useful to develop a common vocabulary that connects the normative, technological, and empirical difficulties of bias reduction for our model.

\section*{Acknowledgement}

This work was partially supported by Beijing Natural Science Foundation under Grant No. 4222027,  National Natural Science Foundation of China under Grant No. 61872369, Beijing Outstanding Young Scientist Program under Grant No. BJJWZYJH012019100020098, 
and the Outstanding Innovative Talents Cultivation Funded Programs 2021 of Renmin University of China. Xin Zhao is the corresponding author.

\bibliography{anthology}
\bibliographystyle{acl_natbib}

\newpage
\appendix

\section*{Appendix}
We provide some experiment-related information as supplementary materials. The appendix is organized into three sections:
\begin{itemize}
	\item Statistics of each dataset are presented in Appendix~\ref{app-dataset};
	\item Training settings of baselines and our model PTG are presented in Appendix~\ref{app-configuration}.
\end{itemize}

\section{Statistics of Datasets} \label{app-dataset}
The detailed information of the dataset for each task is listed in Table~\ref{tab:dataset}, including summarization (CNN/Daily Mail, XSum, MSNews, Multi-News and NEWSROOM), question generation (SQuAD), style transfer (Wiki Neutrality), text paraphrase (Quora), dialog (PersonaChat, TopicalChat, DailyDialog, DSTC7-AVSD and MultiWOZ) and story generation (WritingPrompts). These datasets are utilized under MIT license.

\section{Configuration of Models} \label{app-configuration}

The learning rate of other baselines is set to $3 \times 10^{-5}$, which is the same as our backbone BART. The other settings of baselines and our model are set the same for fair comparison. And we do not utilize special tricks such as label smoothing, warm-up learning rate and length penalty. We apply the Adam optimizer and set $\beta_1=0.9$, $\beta_2=0.98$, $\epsilon=1 \times 10^{-6}$. We set the accumulated batch size of each model to $96$ using accumulated gradients. Furthermore, we use the model with the best performance on validation set for generation. During inference, we apply the beam search method with a beam size of $5$ and a no repeat ngram size of $3$. We train our models using NVIDIA A100 GPUs and PyTorch 1.9.0 upon Ubuntu 20.04.2 LTS.

\begin{table*}[h]
	\centering
	\begin{tabular}{lrrrrr}
		\toprule
		Dataset & \multicolumn{1}{c}{\#Train} & \multicolumn{1}{c}{\#Valid} & \multicolumn{1}{c}{\#Test} & \multicolumn{1}{c}{\#Input} & \multicolumn{1}{c}{\#Output} \\ \midrule
		CNN/Daily Mail & 287113 & 13368 & 11490 & 790.2 & 58.4 \\
		Xsum & 204017 & 11327 & 11333 & 358.8 & 21.1 \\
		MSNews & 136082 & 7496 & 7562 & 311.6 & 24.8 \\
		Multi-News & 44972 & 5622 & 5622 & 2291.9 & 263.1 \\
		NEWSROOM & 995040 & 108837 & 108862 & 658.5 & 26.7 \\
		SQuAD & 75722 & 10570 & 11877 & 148.3 & 11.6 \\
		Wiki Neutrality & 145197 & 18149 & 18150 & 29.1 & 27.3 \\
		Quora & 119410 & 14927 & 14926 & 9.8 & 9.9 \\
		PersonaChat & 122499 & 14602 & 14056 & 122.1 & 11.9 \\
		TopicalChat & 179750 & 11142 & 11221 & 216.6 & 20.3 \\
		DailyDialog & 76052 & 7069 & 6740 & 68.4 & 13.9 \\
		DSTC7-AVSD & 145521 & 33953 & 11780 & 90.7 & 9.5 \\
		MultiWOZ & 105115 & 13748 & 13744 & 110.7 & 13.2 \\
		WritingPrompts & 67765 & 3952 & 3784 & 25.7 & 232.3 \\ \bottomrule
	\end{tabular}
	\caption{Statistics of our datasets after preprocessing. \#Train, \#Valid and \#Test denote the number of examples in training, valid and test datasets, respectively. \#Input and \#Output denote the average number of tokens in the input text and output text.}
	\label{tab:dataset}
\end{table*}

\end{document}